\documentclass[preprint,5p,twocolumn,times,authoryear]{elsarticle}
\usepackage{amsmath,amssymb,amsfonts}
\usepackage{colortbl}
\usepackage{algorithmic}
\usepackage{graphicx}
\usepackage{algorithm,algorithmic}
\usepackage{multirow}
\usepackage{diagbox}
\usepackage{booktabs}
\usepackage[dvipsnames]{xcolor}
\usepackage[colorlinks]{hyperref}

\newtheorem{theorem}{Theorem}

\date{}

\begin{document}

\begin{frontmatter}

\author[label1,label2]{Zhiqiang Shen}
\author[label1,label2]{Peng Cao\corref{cor1}}
\author[label1,label2]{Junming Su}
\author[label1,label2]{Jinzhu Yang}
\author[label3]{Osmar R. Zaiane}
\cortext[cor1]{Corresponding author. 
School of Computer Science and Engineering, Northeastern University, Shenyang, China.
E-mail: \href{mailto:xxszqyy@gmail.com}{xxszqyy@gmail.com}; \href{mailto:caopeng@mail.neu.edu.cn}{caopeng@mail.neu.edu.cn}.}
\affiliation[label1]{organization={School of Computer Science and Engineering, Northeastern University},
            city={Shenyang},
            postcode={110819},
            country={China}}

\affiliation[label2]{organization={Key Laboratory of Intelligent Computing in Medical Image, Ministry of Education},
            city={Shenyang},
            postcode={110819},
            country={China}}

\affiliation[label3]{organization={Alberta Machine Intelligence Institute, University of Alberta},
            city={Edmonton},
            state={Alberta},
            country={Canada}}

\title{Adaptive Mix for Semi-Supervised Medical Image Segmentation}

\begin{abstract}
Mix-up is a key technique for consistency regularization-based semi-supervised learning methods, blending two or more images to generate strong-perturbed samples for strong-weak pseudo supervision. 
Existing mix-up operations are performed either randomly or with predefined fixed rules, such as replacing low-confidence patches with high-confidence ones. The former lacks control over the perturbation degree, leading to overfitting on randomly perturbed samples, while the latter tends to generate images with trivial perturbations, both of which limit the effectiveness of consistency regularization.
This paper aims to answer the following question: How can image mix-up perturbation be adaptively performed during training?
To this end, we propose an \textbf{Ada}ptive \textbf{Mix} algorithm (AdaMix) for image mix-up in a self-paced learning manner. 
Given that, in general, a model's performance gradually improves during training, AdaMix is equipped with a self-paced curriculum that, in the initial training stage, provides relatively simple perturbed samples and then gradually increases the difficulty of perturbed images by adaptively controlling the perturbation degree based on the model's learning state estimated by a self-paced regularizer.
We develop three frameworks with our AdaMix, \emph{i.e.}, AdaMix-ST, AdaMix-MT, and AdaMix-CT, for semi-supervised medical image segmentation. 
Extensive experiments on three public datasets, including both 2D and 3D modalities, show that the proposed frameworks are capable of achieving superior performance. For example, compared with the state-of-the-art, AdaMix-CT achieves relative improvements of 2.62\% in Dice similarity coefficient and 48.25\% in average surface distance on the ACDC dataset with 10\% labeled data.
The results demonstrate that mix-up operations with dynamically adjusted perturbation strength based on the segmentation model's state can significantly enhance the effectiveness of consistency regularization.
The code will be released at \href{https://github.com/Senyh/AdaMix}{\textit{\texttt{https://github.com/Senyh/AdaMix}}}.
\end{abstract}


\end{frontmatter}

\section{Introduction}
\label{sec:introduction}

\begin{figure}[!t]
\centerline{\includegraphics[width=\linewidth]{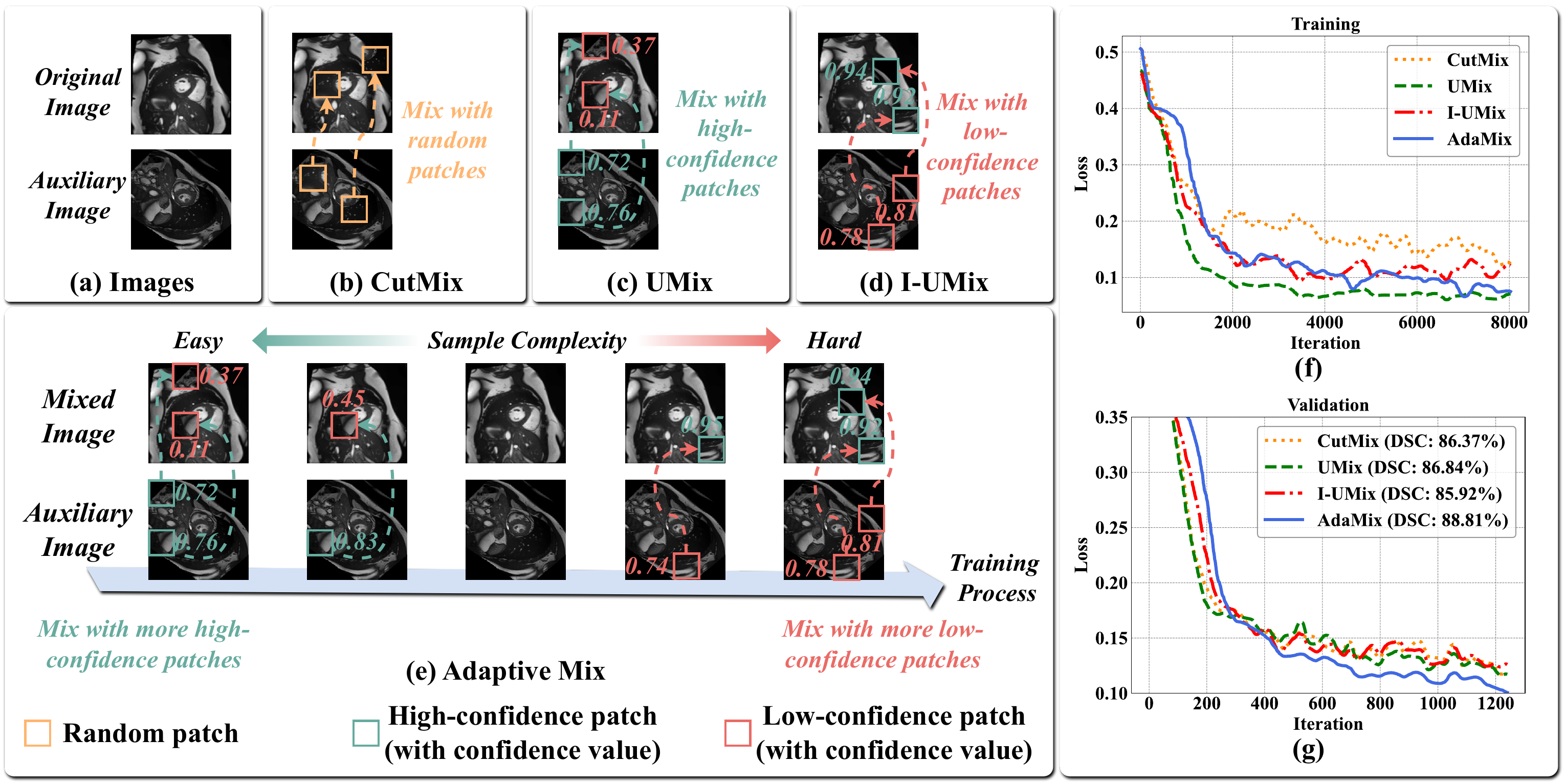}}
\caption{Comparison among CutMix, UMix~\citep{shen2023co}, inverse UMix, and our AdaMix.
(a) Original and auxiliary images;
(b) CutMix randomly replaces patches of the original image with patches from the auxiliary image, with an uncontrollable perturbation degree;
(c) UMix replaces low-confidence patches in the original image with high-confidence ones from the auxiliary image, often generating perturbed examples with trivial perturbations;
(d) I-UMix mixes high-confidence patches in the original image with low-confidence ones from the auxiliary image, yielding overly strong-perturbed images;
(e) Our AdaMix synthesizes images with more high-confidence regions in the initial training stage and then gradually increases the perturbation degree to enhance the complexity of the perturbed images as training progresses;
(f) The unsupervised learning loss curves;
(g) The validation loss curves (and the Dice Similarity Coefficients on the ACDC \textit{test} set).
}
\label{fig:mix_intro}
\end{figure}

Medical image segmentation for delineating tissues, organs, and lesions from different imaging modalities is critical for computer-aided clinical applications, such as prognosis and preoperative planning. Deep learning has greatly advanced the progress of medical image segmentation, relying on its powerful representation capability and large amounts of labeled data~\citep{ronneberger2015u,milletari2016v,chen2021transunet,wang2022uctransnet,wu2024medsegdiff}. 
However, obtaining pixel-level annotations is laborious and expensive in medical images, as it requires expert knowledge. 
Semi-supervised learning (SSL) has the potential to alleviate the label scarcity problem by leveraging a large amount of unlabeled data~\citep{van2020survey}. 

Consistency regularization, one of the most widely used SSL techniques, encourages model predictions to be invariant to input perturbations, aiming to explore supervision information from unlabeled data~\citep{rasmus2015semi,laine2016temporal,tarvainen2017mean}.
Building upon the idea of consistency regularization, strong-weak pseudo supervision~\citep{sohn2020fixmatch} enforces consistency between the predicted pseudo labels of strongly and weakly perturbed unlabeled images, thereby sufficiently extracting supervisory signals from unlabeled data~\citep{sohn2020fixmatch,chen2023softmatch,chen2021semi,wu2022mutual,shen2023co,wang2023mcf}. 
The key challenge of strong-weak pseudo supervision is determining appropriate strong perturbation strategies for generating strong-perturbed images. Improper perturbations cannot match the training model's ability to capture supervision signals from unlabeled data, misleading the training process~\citep{shen2023co}. 
Recently, CutMix\footnote{CutMix is formulated as $\tilde{x} = x_o \times m + x_a \times (1 - m)$ and $\tilde{y} = y_o \times m + y_a \times (1 - m)$, where $x_o$ / $x_a$ denotes an original/auxiliary image,  $y_o$ / $y_a$ refers to the corresponding (pseudo) label, and $m$ is a randomly generated mask. $\tilde{x}$ / $\tilde{y}$ represents the mixed image / (pseudo) label.}\citep{yun2019cutmix}, incorporated with image color space transformation, has been widely used to form the strong perturbations~\citep{french2019semi,chen2021semi,yang2022st++,yang2023unimatch}.
As illustrated in Fig.~\ref{fig:mix_intro}(b), CutMix randomly mixes image patches between an original image and an auxiliary image to synthesize a strong-perturbed example. 
However, such a random mix operation with an uncontrollable perturbation degree leads to an oscillating and unstable consistency regularization process [Fig.~\ref{fig:mix_intro}(f)], where the model is prone to over-fit to the randomly perturbed examples, resulting in unsatisfactory generalization on the validation set [Fig.~\ref{fig:mix_intro}(g)].
To address the instability of the random mix-up operation, our previous work, UMix~\citep{shen2023co} [Fig.~\ref{fig:mix_intro}(c)] synthesizes strong-perturbed images with more high-confidence regions, enabling the model to produce high-quality predictions for these samples. 
As shown in Fig.~\ref{fig:mix_intro}(f-g), the model trained with UMix images demonstrates smoother and lower training losses but achieves inferior validation and test performance due to the limited effectiveness of consistency regularization. 
Furthermore, we investigated the inverse UMix (I-UMix) strategy [Fig.~\ref{fig:mix_intro}(d)]. 
Since this operation yields new samples with overly strong perturbations, the model shows severe oscillation in the training loss and obtains poor validation performance [Fig.~\ref{fig:mix_intro}(f-g)].
It can be observed that random perturbations tend to be uncontrollable, while mix-up operations with fixed rules result in trivial or over-strong perturbations, both of which limit the effectiveness of consistency regularization.
Therefore, developing an adaptive perturbation/augmentation strategy for semi-supervised medical image segmentation is highly desirable.

Conceptually, a model's performance should vary dynamically as the number of training epochs increases; thereby, an appropriate mix-up operation should also dynamically adjust the perturbation degree based on the state of the segmentation model's capability.
Given this assumption, we propose \textbf{Ada}ptive \textbf{Mix} (AdaMix), a novel \textbf{\textit{dynamical and learnable}} perturbation algorithm that adaptively performs image mix-up based on a model's learning state, for better achieving consistency regularization in semi-supervised medical image segmentation. 
Our key idea lies in introducing the concept of self-paced learning~\citep{kumar2010self,jiang2014self,ma2017self} into the design of the mix-up algorithm and providing more reasonable perturbed images with gradually increasing sample complexity for strong-weak pseudo supervision, as depicted in Fig.~\ref{fig:mix_intro}(e).
The mix-up operation with varying degrees of perturbation leads to the generation of perturbed images with different levels of sample complexity.
An image that contains more high-confidence regions is considered an easier sample by a segmentation model, as it tends to produce more accurate predictions from such an image; in contrast, mixing with more high-confidence patches results in a harder example (with a stronger perturbation strength).
More specifically, based on the training loss, AdaMix generates 1) a self-paced mask for determining whether an image is mixed with high-confidence patches or low-confidence ones, and simultaneously provides 2) a self-paced weight for determining the number of mix-up patches.
One can observe from Fig.~\ref{fig:mix_intro}(f-g) that, compared with the random and fixed counterparts (\emph{i.e.}, CutMix~\citep{yun2019cutmix} and UMix~\citep{shen2023co}),  AdaMix is capable of maintaining stable converging ability during the training process and achieving better generalization in the validation and test phase.
We empirically demonstrate that AdaMix can be seamlessly integrated into self-training, mean-teacher, and co-training (\emph{i.e.}, FixMatch~\citep{sohn2020fixmatch}, MT~\citep{tarvainen2017mean}, and CPS~\citep{chen2021semi}) SSL paradigms, resulting in AdaMix-ST, AdaMix-MT, and AdaMix-CT frameworks, respectively, which achieve state-of-the-art performance in semi-supervised medical image segmentation tasks.
We evaluated the proposed frameworks on three public medical image segmentation datasets, involving 2D and 3D modalities. The results demonstrate the effectiveness of each component of our method and its superiority over the state-of-the-art. 

In summary, our main contributions are:
\begin{itemize}
\item \textbf{New Perspective}: We identify the limitations in existing mix-up methods from a dynamic model training perspective: they perform image mix-up either in a random manner or based on pre-defined fixed rules, resulting in perturbed images with uncontrollable, trivial, or over-strong perturbation degrees, which seriously undermine the effectiveness of consistency regularization. 
We suggest that an adaptive mix-up scheme with awareness of the model's learning state is critical for consistency regularization.
\item \textbf{New method}: We propose a novel algorithm, AdaMix, for image mix-up in a self-paced manner. To the best of our knowledge, this is the first work to adaptively perform mix-up perturbation based on a model's state, where the segmentation model and the AdaMix algorithm collaboratively learn from each other. Furthermore, AdaMix is highly flexible and can be seamlessly integrated into self-training, mean-teacher, and co-training SSL paradigms.
\item \textbf{New findings}: 
We provide new insights: incorporating high-confidence patches into images generates simpler samples, which benefits the segmentation model during the initial training stage. 
In contrast, the low-confidence regions, which typically appear at segmentation boundaries, force the segmentation model to focus on these hard regions and enhance the model’s discriminative capability when mixed into images in the later training stages. 
We also reveal that perturbation strategies are more important than learning paradigms for consistency regularization. 
\end{itemize}

\section{Related Work}
\label{sec:related_work}

\subsection{Semi-Supervised Learning}
\label{subsec:semi-supervised_learning}
Semi-supervised learning (SSL) is an effective learning strategy that addresses the issue of limited labeled data by utilizing a large amount of unlabeled images~\citep{van2020survey}. 
Based on the design of the regularization term to leverage unlabeled data, many SSL algorithms have been proposed, such as adversarial learning~\citep{springenberg2015unsupervised,miyato2018virtual}, pseudo labeling~\citep{lee2013pseudo}, and consistency regularization~\citep{rasmus2015semi,laine2016temporal}. 
Recently, state-of-the-art SSL approaches have incorporated the idea of strong-weak pseudo supervision~\citep{sohn2020fixmatch} as the primary strategy of their frameworks. According to the learning paradigms, these approaches typically fall into three branches: self-training-based 
\citep{berthelot2019mixmatch,berthelot2019remixmatch,sohn2020fixmatch,zhang2021flexmatch,wang2023freematch,chen2023softmatch}, 
mean-teacher-based~\citep{tarvainen2017mean} and co-training-based frameworks~\citep{ke2019dual,li2023multi}. 

Similar to semi-supervised image classification tasks, the line of semi-supervised segmentation methods also can be divided into three branches: self-training-based~\citep{zou2021pseudoseg,yuan2021simple,yang2022st++,yang2023unimatch}, 
mean-teacher-based~\citep{yu2019uamt,hu2021semi,xu2022semi,liu2022perturbed},
and co-training-based methods 
\citep{chen2021semi,wang2023conflict,luo2021semi,wu2021semi,wu2022mutual,wang2023mcf,shen2023co}.
The self-training paradigm involves a single model that generates pseudo labels for training the model per se~\citep{sohn2020fixmatch}.
Mean-Teacher~\citep{tarvainen2017mean} can be considered an intermediate approach between self-training and co-training. It involves a teacher and a student model, where pseudo-supervision is conducted from the teacher to the student.
A representative co-training paradigm, presented in CPS~\citep{chen2021semi}, comprises two (student) models that are cross-supervised by each other.
We find that perturbation strategies are more important than a specific learning paradigm. 
We propose a novel image mix-up algorithm that can be seamlessly applied to these paradigms and achieves state-of-the-art performance in semi-supervised medical image segmentation.

\subsection{Mix-up Augmentation}
\label{subsec:mixup}
Mix-up operations can be regarded as a line of data augmentation approaches that blend two or more images and their corresponding labels to synthesize a new example.
A series of mix-up strategies were initially proposed in supervised learning to generate new samples via randomly mixing images in image-level~\citep{zhang2017mixup}, feature-level~\citep{verma2019manifold}, and region-level~\citep{devries2017improved,yun2019cutmix,liu2022automix}.
Furthermore, additional information, such as saliency maps~\citep{kim2020puzzle,kim2021comixup,uddin2021saliencymix,kang2023guidedmixup}, was incorporated into the mix-up algorithms.
These methods extend training data distributions and impose strong regularization during model training, implicitly controlling model complexity and enhancing model robustness.

Recently, mix-up algorithms have been utilized to introduce strong perturbations for consistency regularization in semi-supervised learning frameworks for image recognition~\citep{berthelot2019mixmatch,berthelot2019remixmatch,sohn2020fixmatch,verma2022interpolation} and semantic segmentation~\citep{french2019semi,zou2020pseudoseg,chen2021semi,yang2023revisiting,shen2023co,chi2024adaptive}.
Specifically, they are performed in fully random manners (\emph{e.g.}, Mixup, Cutout, or CutMix) ~\citep{berthelot2019mixmatch,berthelot2019remixmatch,sohn2020fixmatch,french2019semi,zou2020pseudoseg,chen2021semi,verma2022interpolation},
or according to predefined fixed rules (\emph{e.g.}, UMix that replaces low-confidence patches with high-confidence ones guided by predictive uncertainty/confidence)~\citep{shen2023co,chi2024adaptive}.
However, the strength of the random perturbation is uncontrollable, leading to overfitting to randomly perturbed samples, while the degree of the fixed mix-up operations is trivial or over-strong. Both cases generate improper perturbed images limiting the effectiveness of consistency regularization. 
To address this issue, an adaptive mix-up perturbation scheme is highly desirable for effective consistency regularization in semi-supervised medical image segmentation.

\begin{figure*}[!t]
\centerline{\includegraphics[width=\textwidth]{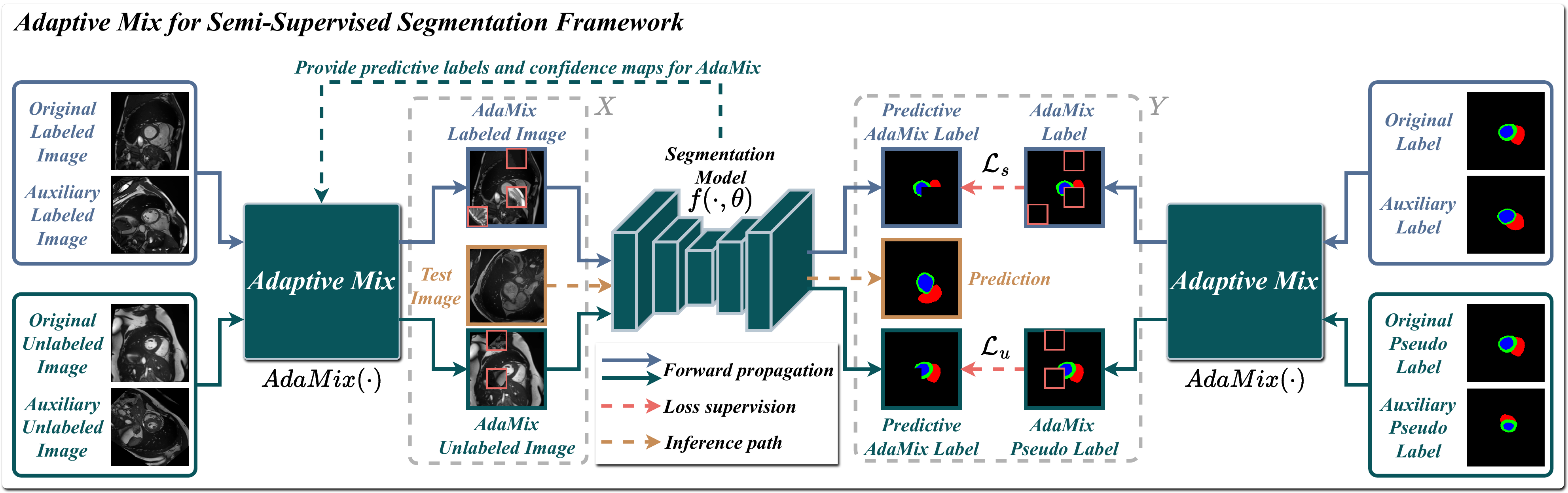}}
\caption{Overview of the proposed Adaptive Mix framework for semi-supervised medical image segmentation. It includes our Adaptive Mix (AdaMix) algorithm for generating perturbed images and a semi-supervised learning paradigm for providing pseudo labels and conducting consistency regularization. AdaMix can be seamlessly applied to the self-training, mean-teacher, and co-training paradigms, resulting in  AdaMix-ST, AdaMix-MT, and AdaMix-CT frameworks, respectively.
}
\label{fig:adamix_framework}
\end{figure*}

\subsection{Medical Image Segmentation}
\label{subsec:medical_image_segmentation}
Deep learning-based methods have achieved remarkable progress in medical image segmentation tasks~\citep{ronneberger2015u,cciccek20163d,milletari2016v}.
Specifically, building upon fully convolutional networks~\citep{long2015fully}, U-Net and its variants~\citep{ronneberger2015u,cciccek20163d,milletari2016v,zhou2018unet++,oktay2018attention,isensee2021nnu} have been the mainstream baseline segmentation models widely adopted in further studies, such as semi-supervised segmentation and related tasks~\citep{yu2019uamt,wu2021semi,shen2023co}. 
Furthermore, to address the issues of multi-scale heterogeneous organs and lesions in medical imaging, numerous transformer-based models ~\citep{chen2021transunet,valanarasu2021medical,cao2022swin,wang2022uctransnet,wang2024narrowing} have been applied to model long-range dependencies among segmentation targets. 
Meanwhile, diffusion models-based approaches have recently been developed to improve the robustness of segmentation models against noise and artifacts of medical images~\citep{wu2024medsegdiff,wu2024medsegdiffv2}. 
More recently, Segment Anything Models~\citep{ma2024segment,wei2024medsam} emerged as a foundation model aiming at universal medical image segmentation across a wide spectrum of tasks.
Although these existing methods show promising results, their performance relies heavily on large-scale datasets with numerous pixel-level annotations, which are time-consuming and expertise-demanding in specific clinical applications, \emph{e.g.} assessment of cardiac ejection fraction and diagnosis of skin melanoma.
In contrast, our AdaMix algorithm can function as a plug-and-play, model-agnostic module, enabling consistency regularization on these fully supervised models to leverage unlabeled data for alleviating the annotation burden.

\section{Method}
\label{sec:method}
\subsection{Problem statement}
\label{subsec:problem_statement}
Before delving into the proposed method, we first provide the notations that will be used subsequently. The training set $\mathcal{D} = \{\mathcal{D}^L, \mathcal{D}^U\}$ contains a labeled set $\mathcal{D}^L = \{(x^l_i, y^l_i)_{i=1}^{N^L}\}$ and an unlabeled set $\mathcal{D}^U = \{(x^u_j)_{j=1}^{N^U}\}$, where $x^l_i$/$x^u_j$ denotes the $i^{th}$/$j^{th}$ labeled/unlabeled image, $y^l_i$ is the ground truth for the labeled image, and $N^L$ and $N^U$ ($N^U >> N^L$) are the numbers of labeled and unlabeled samples. Given the training data $\mathcal{D}$, semi-supervised semantic segmentation aims to learn a model $f(\cdot; \theta)$ that performs well on unseen test sets $\mathcal{T}$.

\subsection{Overview}
\label{subsec:overview}
\begin{figure*}[!t]
\centerline{\includegraphics[width=\textwidth]{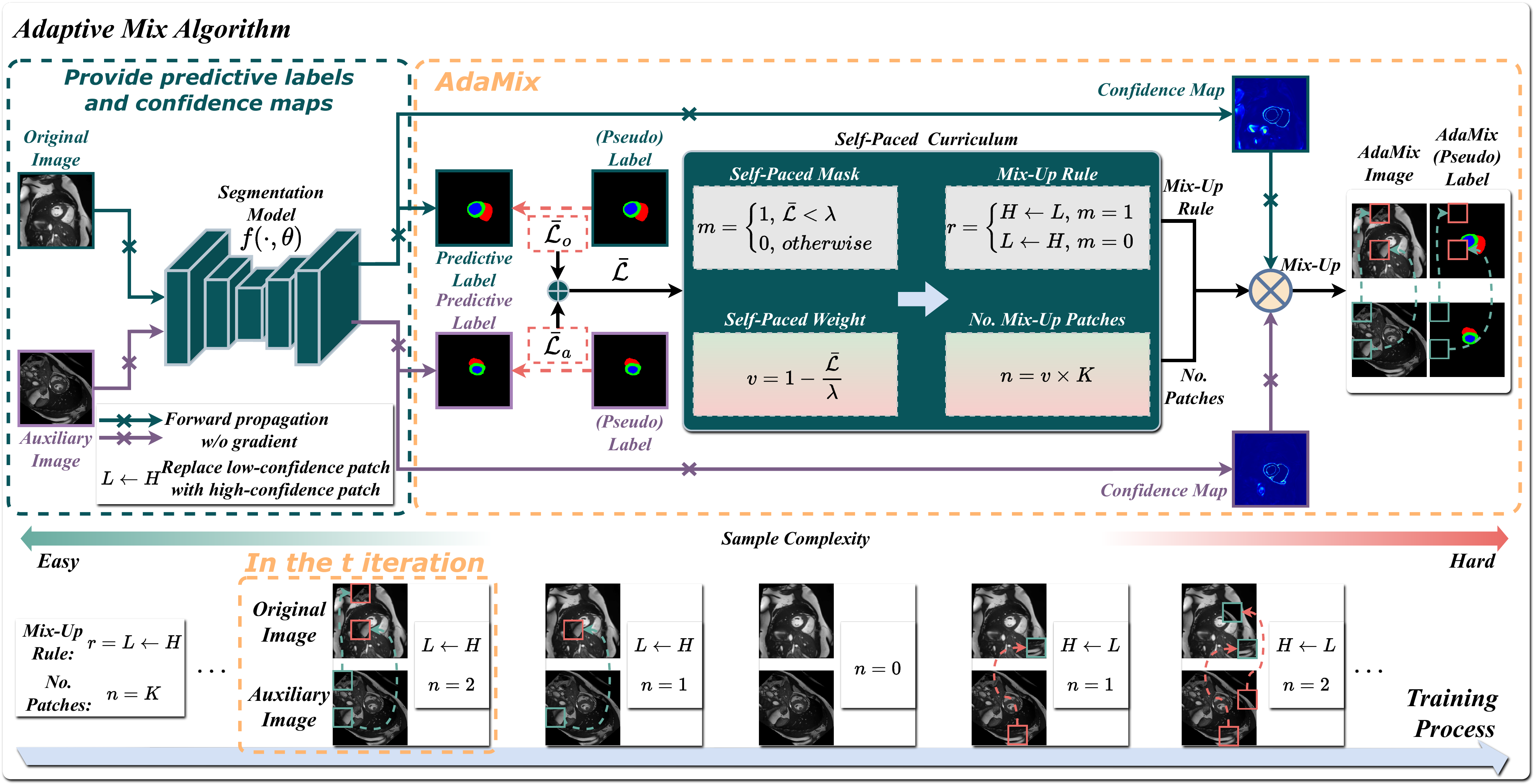}}
\caption{Schematic diagram of the proposed Adaptive Mix algorithm (AdaMix). It performs image mix-up perturbation in a self-paced manner, synthesizing perturbed samples from easy to hard during training based on the model's learning state.
}
\label{fig:adamix_algorithm}
\end{figure*}

As discussed in Section~\ref{sec:introduction}, the random region mix-up~\citep{yun2019cutmix} does not allow for adaptively controlling perturbation degree, leading to the learned model overfitting to randomly perturbed examples. Meanwhile, the confidence-guided mix~\citep{shen2023co}, which utilizes a fixed mix-up rule that replaces low-confidence patches with high-confidence ones, produces mixed images with trivial perturbation degrees, thereby limiting the effect of consistency regularization. In contrast, replacing high-confidence regions with low-confidence regions may generate new samples with overly strong perturbations, making it difficult for the model to adapt to these samples. 
We pinpoint the question: \textit{How can image mix-up augmentation be adaptively performed during training?} 
To this end, we propose Adaptive Mix (AdaMix), a novel augmentation algorithm for adaptive image mix-up based on a model's learning state. 
Building upon AdaMix, we devise an Adaptive Mix framework, where AdaMix acts as a plug-and-play module and can be seamlessly integrated into the self-training, mean-teacher, and co-training paradigms for semi-supervised medical image segmentation. 
In the following, we first specify the Adaptive Mix framework [Fig.~\ref{fig:adamix_framework}] and then elaborate on the Adaptive Mix algorithm [Fig.~\ref{fig:adamix_algorithm}].

\definecolor{commentcolor}{RGB}{63,136,196}  
\newcommand{\PyComment}[1]{\ttfamily\textcolor{commentcolor}{\# #1}}
\newcommand{\PyCode}[1]{\ttfamily\textcolor{black}{#1}}
\begin{algorithm}[!t]
    \caption{AdaMix-ST}
    \label{alg:adamix_st}
    \PyComment{Take self-training as an example}\\
    \textbf{Input}: {$\mathcal{D}^L = \{(x^l_i, y^l_i)_{i=1}^{N^L}\}$, $\mathcal{D}^U = \{(x^u_j)_{j=1}^{N^U}\}$}\\
    \textbf{Output}: {Trained segmentation model $f(\cdot; \theta)$}
    \begin{algorithmic}[1] 
        \FOR{each iteration}
                \STATE {\PyComment{Forward propagation (w/o gradient) to obtain pseudo labels for unlabeled original/auxiliary images $x^u_j/ x^u_{b}$ ($j \ne b$)}}
                \STATE {$\bar{y}^u_j \gets f(x^u_j, \theta)$ and $\bar{y}^u_{b} \gets f(x^u_{b}, \theta)$}
                \STATE {\PyComment{Perform AdaMix to obtain augmented labeled and unlabeled images}}
                \STATE {$\tilde{x}^l_i, \tilde{y}^l_i \gets \mathrm{AdaMix}(x^l_i, x^l_{a}, y^l_i, y^l_{a}, f(\cdot, \theta))$}
                \STATE {$\tilde{x}^u_j, \tilde{y}^u_j \gets \mathrm{AdaMix}(x^u_j, x^u_{b}, \bar{y}^u_j, \bar{y}^u_{a}, f(\cdot, \theta))$}
                \STATE {\PyComment{Forward propagation to obtain predictions for labeled and unlabeled perturbed images $\tilde{x}^l_i$ and $\tilde{x}^u_j$}}
                \STATE {$\hat{y}^l_i \gets f(\tilde{x}^l_i, \theta)$ and $\hat{y}^u_j \gets f(\tilde{x}^u_j, \theta)$}
                \STATE {\PyComment{Loss supervision by Eq.~\eqref{eq:ssl}}}
                \STATE {Calculate $\mathcal{L}_{s}$ based on $\hat{y}^l_i$, $\tilde{y}^l_i$} 
                \STATE {Calculate $\mathcal{L}_{u}$ based on $\hat{y}^u_j$, $\tilde{y}^u_j$}
                \STATE {Back-Propagate $\mathcal{L} = \mathcal{L}_{s} + \mathcal{L}_{u}$}
                \STATE {Update $f(\cdot; \theta)$}
        \ENDFOR
        \STATE \textbf{return} $f(\cdot; \theta)$
    \end{algorithmic}
\end{algorithm}

\subsection{Adaptive Mix Framework}
\label{subsec:self-paced_mix_framework}
As illustrated in Fig.~\ref{fig:adamix_framework}, the AdaMix framework includes two components: 1) the AdaMix algorithm for generating perturbed examples and 2) a semi-supervised learning scheme for conducting supervision on both labeled and unlabeled images.
According to the approach of strong-weak pseudo supervision~\citep{sohn2020fixmatch}, semi-supervised learning can be realized through three paradigms: 1) self-training, 2) mean-teacher, and 3) co-training, in which FixMatch~\citep{sohn2020fixmatch}, Mean-Teacher~\citep{tarvainen2017mean}, and CPS~\citep{chen2021semi} are the corresponding representative frameworks, respectively. 

Formally, these paradigms can be formulated as follows.
Given $(x^l_i, y^l_i) \in \mathcal{D}^L$, $x^u_j \in \mathcal{D}^U$ and its corresponding pseudo label $\bar{y}^u_j$ with the confidence map $\bar{p}^u_j$, the semi-supervised learning objective can be defined as:
\begin{equation}
J = \mathop{\min}\limits_{\theta} \mathbb{E}_{\mathcal{D}}[\underbrace{\mathcal{L}_{seg}(f(x^l_i, \theta), y^l_i)}_{\mathcal{L}_{s}} + \underbrace{\mathbb{I}_{\{\bar{p}^u_j \geq \tau\}}(\bar{p}^u_j)\mathcal{L}_{seg}(f(\Gamma(x^u_j), \theta), \bar{y}^u_j)}_{\mathcal{L}_{u}}]\\
\label{eq:ssl}
\end{equation}
where the first and second terms denote the supervised loss $\mathcal{L}_s$ and the unsupervised loss $\mathcal{L}_u$, respectively. $\mathcal{L}_{seg}(\cdot, \cdot)$ is an image segmentation criterion, $f(\cdot, \theta)$ denotes a segmentation model with learnable parameters $\theta$, and $\tau$ refers to a confidence threshold above which a pixel-level pseudo label is retained for pseudo supervision. 
$\Gamma(\cdot)$ represents a strong perturbation/augmentation, which is the key to the efficacy of consistency regularization.

\textbf{\textit{Self-Training}} involves a single model $f(\cdot, \theta)$, which generates pseudo labels from unlabeled data for training of the model itself. A pseudo label is generated by: $\bar{y}^u_j = \arg\max{\mathrm{SoftMax}(f(\Gamma'(x^u_j), \theta))}$, where $\mathrm{SoftMax} (\cdot)$ denotes the SoftMax function that converts predictive logits into probabilities and $\Gamma'(\cdot)$ represents a weak perturbation (\emph{w.r.t.} the strong perturbation $\Gamma(\cdot)$) \footnote{Following the idea of strong-weak pseudo supervision~\citep{sohn2020fixmatch}, in practice, $\Gamma'(\cdot)$ refers to "weak" perturbations, generally consisting of only geometric augmentations, while $\Gamma(\cdot)$ refers to "strong" augmentations, which include geometric and pixel intensity augmentations, as well as mix-up operations. For brevity, these notations are omitted in the following text.}.

\textbf{\textit{Mean-Teacher}} includes a student model $f(\cdot, \theta)$ and a teacher $f(\cdot, \bar{\theta})$ model, where the parameters $\bar{\theta}$ of the teacher model is updated by an exponential moving average (EMA) of the student's parameters $\theta$ at each iteration $t$: $\bar{\theta}_t = \alpha \bar{\theta}_{t-1} + (1-\alpha) \theta_t$. Pseudo labels are generated by the teacher model: $\bar{y}^u_j = \arg\max{\mathrm{SoftMax}(f(\Gamma'(x^u_j), \bar{\theta}))}$.

\textbf{\textit{Co-Training}} involves two (student) models, $f(\cdot, \theta_1)$ and $f(\cdot, \theta_2)$, which are cross-supervised from each other on unlabeled data. The pseudo labels yielded by one model are leveraged to supervise another model, \emph{i.e.}, $\bar{y}^u_{j[1/2]} = \arg\max{\mathrm{SoftMax}(f(\Gamma'(x^u_j), \theta_{[1/2]}))}$.

Building upon these paradigms and the strong-weak pseudo supervision, we develop three frameworks with our AdaMix, \emph{i.e.}, AdaMix-ST, AdaMix-MT, and AdaMix-CT, for semi-supervised medical image segmentation. 
These instantiated frameworks employ Eq.~\eqref{eq:ssl} as their learning objective and exploit AdaMix as the strong perturbation $\Gamma(\cdot)$ for consistency regularization. The training procedure of these methods is detailed in Algorithm~\ref{alg:adamix_st} (Note that, without loss of generality, we take AdaMix-ST as an example for illustration convenience).
We will show in Section~\ref{sec:experiments_and_results} that the proposed frameworks achieve state-of-the-art performance on various medical image segmentation tasks.

\begin{algorithm}[!t]
    \caption{AdaMix}
    \label{alg:adamix}
    \PyComment{Take AdaMix on unlabeled data as an example} \\
    \textbf{Input}: {Original image $(x^u_j, \bar{y}^u_j)$ and auxiliary image $(x^u_{b}, \bar{y}^u_{b})$ ($j\ne b$)}\\
    \textbf{Output}: {AdaMix image $\tilde{x}^u_j, \tilde{y}^u_j$}
    \begin{algorithmic}[1] 
            \STATE {\PyComment{Forward propagation (w/o gradient) to obtain predictions and confidence maps}}
            \STATE {$\hat{y}^u_j, \hat{p}^u_j \gets f(x^u_j, \theta)$ and $\hat{y}^u_{b}, \hat{p}^u_{b} \gets f(x^u_{b}, \theta)$}
            \STATE {\PyComment{Self-Paced Curriculum}}
            \STATE {Calculate $\bar{\mathcal{L}}$ based on $\hat{y}^u_j, \bar{y}^u_j$ and $\hat{y}^u_{b}, \bar{y}^u_{b}$}
            \STATE {\PyComment{Self-Paced Mask}}
            \STATE {Update $m$ by Eq.~\eqref{eq:self_paced_mask_update} to determine the Mix-Up rule $L \leftarrow H$ or $H \leftarrow L$}
            \STATE {\PyComment{Self-Paced Weight}}
            \STATE {Update $v$ by Eq.~\eqref{eq:self_paced_weight_update} to determine the number of Mix-Up patches $n$}
            \STATE {\PyComment{Mix-up Procedure}}
            \IF{$L \leftarrow H$}
            \STATE {$\tilde{x}^u_j / \tilde{y}^u_j \gets $ Replace top $n$ low-confidence patches in $x^u_j / \bar{y}^u_{j}$ with top $n$ high-confidence patches in $x^u_{b} / \bar{y}^u_{b}$}
            \ELSE
            \STATE {$\tilde{x}^u_j / \tilde{y}^u_j \gets $ Replace top $n$ high-confidence patches in $x^u_j / \bar{y}^u_{j}$ with top $n$ low-confidence patches in $x^u_{b} / \bar{y}^u_{b}$}
            \ENDIF
        \STATE \textbf{return} $\tilde{x}^u_j, \tilde{y}^u_j$
    \end{algorithmic}
\end{algorithm}

\subsection{Adaptive Mix algorithm}
\label{subsec:self-paced_mix_algorithm}
As illustrated in Fig.~\ref{fig:adamix_algorithm}, AdaMix is equipped with a self-paced curriculum that, at each training iteration, provides a self-paced mask and a self-paced weight to determine the mix-up rule and the number of mix-up patches, respectively, based on the model's learning state.
In our framework, AdaMix is applied to both labeled and unlabeled data. For convenience, we illustrate the algorithm details by taking AdaMix on unlabeled data as an example in the following. A detailed description of AdaMix on unlabeled data is provided in Algorithm~\ref{alg:adamix}.  Note that the only difference between AdaMix on labeled and unlabeled data is that the former uses ground truth segmentation labels, while the latter utilizes pseudo labels.

Given an unlabeled original image $x^u_j$, an unlabeled auxiliary image $x^u_{b}$, and their corresponding pseudo labels $\bar{y}^u_{j}, \bar{y}^u_{b}$, we first obtain their confidence maps by $\hat{p}^u_{j} = \max(\mathrm{SoftMax}(f(x^u_j, \theta)))$ and $\hat{p}^u_{b} = \max(\mathrm{SoftMax}(f(x^u_{b}, \theta)))$, respectively.
Meanwhile, we calculate the self-paced proxy loss $\bar{\mathcal{L}} = \bar{\mathcal{L}}_o + \bar{\mathcal{L}}_a$ as an indicator of the model's state, where $\bar{\mathcal{L}}_o = \mathcal{L}_{seg}(f(x^u_j, \theta), \bar{y}^u_j)$ and $\bar{\mathcal{L}}_a = \mathcal{L}_{seg}(f(x^u_{b}, \theta), \bar{y}^u_{b})$ denote the segmentation loss for the original image-pseudo label pair and the auxiliary image-pseudo label pair, respectively. 
Based on the self-paced proxy loss $\bar{\mathcal{L}}$, we define the self-paced mask and self-paced weight as follows.

\paragraph{Self-Paced Mask}
Self-paced mask $m \in \{0, 1\}$ is a binary mask used to determine the mix-up rule $r$ per image. Specifically, when $m = 0$, low-confidence patches are replaced with high-confidence patches $L \leftarrow H$; when $m = 1$, high-confidence patches are replaced with low-confidence patches $H \leftarrow L$:
\begin{equation}
   r = \left\{
\begin{aligned}
& H \leftarrow L, if\, m = 1\\
& L \leftarrow H, if\, m = 0
\end{aligned}
\right.
\label{eq:mixup_rule}
\end{equation}
where $H/L$ represents a set of high-confidence/low-confidence patches, and $m$ is obtained by solving the following objective function:
\begin{equation}
    m^*(\bar{\mathcal{L}}, \lambda) = \mathop{\arg\min}\limits_{m\in{0,1}}m\bar{\mathcal{L}} + h(m, \lambda)
    \label{eq:self_paced_mask}
\end{equation}
where $\lambda$ denotes the age parameter for controlling the self-paced learning pace. In general, a model's performance gradually improves as training progresses. Based on this, we employ a time-dependent Gaussian ramp-up function to gradually increase the age parameter, \emph{i.e.}, $\lambda(t) = \exp{[-5(1-\frac{t}{t_m})^2]}$, where $t$ denotes the current iteration and $t_m$ is the maximum iteration in the training process.
$h(m, \lambda)$ refers to a self-paced regularizer~\citep{meng2017theoretical}. 
Considering the binary property of $m$, we employ a hard self-paced regularizer, \emph{i.e.},  $h(m, \lambda) = -\lambda m$, thereby obtaining the corresponding solution for $m$:
\begin{equation}
    m^* = \left\{
\begin{aligned}
& 1, if\, \bar{\mathcal{L}} < \lambda\\
& 0, otherwise
\end{aligned}
\right.
\label{eq:self_paced_mask_update}
\end{equation}

\paragraph{Self-Paced Weight}
Self-paced weight $v \in [0, 1]$ determines the number of mix-up patches $n$ per image, defined as:
$n = v \times K$
where $K$ is a hyperparameter denoting the maximum number of mix-up patches. Similar to the calculation of the self-paced mask, $v$ is obtained by:
\begin{equation}
    v^*(\bar{\mathcal{L}}, \lambda) = \mathop{\arg\min}\limits_{v\in[0,1]}v\bar{\mathcal{L}} + g(v, \lambda)
    \label{eq:self_paced_weight}
\end{equation}
where $g(v, \lambda) = \lambda(\frac{1}{2}v^2 - v)$ is a linear self-paced regularizer. 
By solving Eq.~\eqref{eq:self_paced_weight}, we obtain the solution for $v$:
\begin{equation}
    v^* =  1 - \frac{\bar{\mathcal{L}}}{\lambda}
    \label{eq:self_paced_weight_update}
\end{equation}

Finally, based on the confidence maps $\hat{p}^u_{j}$ and $\hat{p}^u_{b}$, the mix-up rule $r$, and the number of mix-up patches $n$, we perform patch mix-up with patch size $S$ to derive an AdaMix perturbed example $(\tilde{x}^u_j, \tilde{y}^u_j)$. Please refer to Algorithm~\ref{alg:adamix} for a detailed description of the procedure above.

\begin{table*}[!t]
\centering
\caption{Comparison with state-of-the-art methods under the 10\% $|\mathcal{D}|$ and 20\% $|\mathcal{D}|$ labeled data on the ACDC dataset.
Note that all methods adopt U-Net as the backbone.
The best and second-best results are highlighted in {\color{red}\textbf{red}} and {\color{blue}\textbf{blue}}, respectively. 
${}^*$ indicates statistically significant improvements ($p$-value $ < 0.05$ based on a paired t-test) of our AdaMix frameworks over ABD, MT, and UCMT, respectively.}
\resizebox{\textwidth}{!}{
\begin{tabular}{lllll|llllll}
\toprule[1pt]
\multirow{2}{*}{Method} & \multicolumn{4}{c|}{10\% $|\mathcal{D}|$} & \multicolumn{4}{c}{20\% $|\mathcal{D}|$} &\multirow{2}{*}{Perturbation} &\multirow{2}{*}{Framework}\\ \cline{2-9}
& DSC (\%) $\uparrow$ & Jaccard (\%) $\uparrow$ & 95HD $\downarrow$ & ASD $\downarrow$  & DSC (\%) $\uparrow$ & Jaccard (\%) $\uparrow$ & 95HD $\downarrow$ & ASD $\downarrow$  \\ \midrule
URPC~\citep{luo2021urpc}     &    81.77     &      70.85   &     5.04     &     1.41   &    85.07     &      75.61   &     6.26     &      1.77  & Multi-Task & Self-Training \\ 
SASSNet~\citep{li2020shape}     &    84.14       &    74.09     &    5.03       &    1.40       &    87.04       &    78.13     &    7.84       &   2.15 & Multi-Task & Self-Training \\
FixMatch~\citep{sohn2020fixmatch}   & 86.37  & 76.79  & 5.62  & 1.39  & 86.60  & 77.14  & 5.64  & 1.57  & CutMix  & Self-Training \\
ABD~\citep{chi2024adaptive}  & 86.34 & 76.79 & 4.72 & 1.62 & 87.83  & 78.50  & 5.08  & 1.42  & UMix + I-UMix & Self-Training \\    
AdaMix-ST (ours)   &     {\color{blue}\textbf{88.81}}${}^*$      &  {\color{blue}\textbf{80.47}}${}^*$       &    4.38${}^*$      &   1.27${}^*$   &     {\color{blue}\textbf{89.51}}${}^*$      &    {\color{blue}\textbf{81.59}}${}^*$      &    3.79${}^*$      &   1.12  & Adaptive Mix & Self-Training \\ \midrule \midrule 
UA-MT~\citep{yu2019uamt}     &    81.58       &    70.48    &    12.35     &    3.62      &    85.87       &    76.78     &    5.06      &    1.54  & Noise & Mean-Teacher\\ 
MT~\citep{tarvainen2017mean}      &    85.61   &     75.85   &     6.80      &       2.08  &     86.80    &      77.74   &    6.08     &       1.98  & CutMix & Mean-Teacher \\
AdaMix-MT (ours)   &     88.43${}^*$      &  79.91${}^*$       &    {\color{blue}\textbf{3.58}}${}^*$ &   {\color{blue}\textbf{1.18}}${}^*$   &     89.18${}^*$      &    81.12${}^*$      &   {\color{blue}\textbf{3.21}}${}^*$      &   {\color{blue}\textbf{0.96}}${}^*$  & Adaptive Mix & Mean-Teacher  \\  \midrule \midrule
DTC~\citep{luo2021semi}     &    82.71     &      72.14   &     11.31     &     2.99   &    86.28     &      77.03   &     6.14     &      2.11  & Multi-Task & Co-Training \\ 
MC-Net~\citep{wu2021semi}      &     86.34    &      76.82   &     7.08      &       2.08  &     87.83    &     79.14   &     4.94     &      1.52   & Noise & Co-Training \\ 
MC-Net+~\citep{wu2022mutual}      &     87.10    &      78.06   &     6.68      &       2.00  &     88.51    &     80.19   &     5.35     &      1.54  & Noise & Co-Training \\ 
CPS~\citep{chen2021semi}      &    85.84    &      76.15   &     7.06      &       2.17  &     87.07    &      78.11   &    5.38     &       1.82  & CutMix & Co-Training \\  
UCMT~\citep{shen2023co}    &     86.91      &  79.22        &    5.32      &   1.43   &     87.45      &    80.19      &    4.61      &    1.40  & UMix  & Co-Training \\ 
AdaMix-CT (ours)   &     {\color{red}\textbf{89.19}}${}^*$      &  {\color{red}\textbf{81.07}}${}^*$       &    {\color{red}\textbf{2.46}}${}^*$      &   {\color{red}\textbf{0.74}}${}^*$   &     {\color{red}\textbf{89.83}}${}^*$      &    {\color{red}\textbf{81.60}}${}^*$      &    {\color{red}\textbf{2.47}}${}^*$      &   {\color{red}\textbf{0.68}}${}^*$  & Adaptive Mix & Co-Training \\ 
\bottomrule[1pt]
\end{tabular}}
\label{Tab:acdc}
\end{table*} 

\section{Experiments and Results}
\label{sec:experiments_and_results}

\subsection{Datasets}
\label{subsec:datasets}
\textbf{ACDC}\footnote{\href{https://www.creatis.insa-lyon.fr/Challenge/acdc/}{\textit{https://www.creatis.insa-lyon.fr/Challenge/acdc/}}} contains 200 short-axis cine-MRIs from 100 subjects, and the corresponding annotations with the left ventricle, right ventricle, and myocardium labels. Following~\citep{ssl4mis2020}, we split the dataset into the new training, validation, and testing sets, respectively including 70, 10, and 20 patients’ data.

\textbf{LA}\footnote{\href{http://atriaseg2018.cardiacatlas.org/}{\textit{http://atriaseg2018.cardiacatlas.org/}}} consists of 100 3D gadolinium-enhanced magnetic resonance scans and left atrial segmentation ground truths. Following~\citep{yu2019uamt}, we divide the 100 scans into 80 samples for training and 20 samples for evaluation.

\textbf{ISIC}\footnote{\href{https://challenge.isic-archive.com/}{\textit{https://challenge.isic-archive.com/}}} includes 2594 dermoscopy images and the corresponding skin lesion annotations. We divide the entire dataset in a 7:1:2 ratio, resulting in 1815 images for training, 260 images for validation, and 519 images for testing.

\subsection{Implementation Details}
\label{subsec:implementation_details}

\textbf{Experimental environment:} 
Hardware: NVIDIA A40 GPU with 48G GPU memory.
Software: Python 3.8, PyTorch~\citep{paszke2019pytorch} 1.11.0, CUDA 11.3.
We utilize AdamW~\citep{kingma2014adam} as the optimizer with a fixed learning rate of 1e-4. The training time is set to 100 epochs for the ACDC and ISIC datasets and 1000 epochs for the LA dataset.

\textbf{Framework:}
We employ U-Net~\citep{ronneberger2015u}/V-Net~\citep{milletari2016v} as the supervised baseline architecture for 2D/3D image segmentation, respectively. 
We utilize FixMatch~\citep{sohn2020fixmatch}, MT~\citep{tarvainen2017mean}, and CPS~\citep{chen2021semi} as the base frameworks for self-training, mean-teacher, and co-training, respectively. These frameworks are built upon the concept of strong-weak pseudo supervision~\citep{sohn2020fixmatch} to leverage unlabeled data.
We set the confidence threshold $\tau = 0.95$ following FixMatch~\citep{sohn2020fixmatch}.
For the hyperparameters of AdaMix, we set the mix-up patch size $S = 32$ and the maximum number of mix-up patches $K = 16$.
Detailed analysis for these hyperparameters is provided in Section~\ref{subsubsec:hyperparameters}.
In our experiments, we adopt the dice loss~\citep{milletari2016v} as the segmentation criterion.

\textbf{Data:}
In the 2D image segmentation tasks, all images are resized to $256 \times 256$ for inference, while the outputs are recovered to their original sizes for evaluation. For 3D segmentation, we randomly crop $80 \times 112 \times 112 \, (Depth \times Height \times Width)$ patches for training and iteratively crop patches using a sliding window strategy to obtain the final segmentation mask for evaluation. 

\textbf{Evaluation metrics.} Dice similarity coefficient (DSC), Jaccard, average surface distance (ASD), and 95\% Hausdorff distance (95HD) are employed to estimate the segmentation performance in our experiments.

\subsection{Comparison with State of the Arts}
\label{subsec:comparison_with_sota}
We extensively compared our proposed AdaMix frameworks with state-of-the-art semi-supervised learning methods, which can be divided into four groups based on their perturbation strategies: 
1) Multi-task perturbation (using signed distance maps~\citep{li2020shape,luo2021semi} or multi-scale features~\citep{luo2021urpc}): SASSNet~\citep{li2020shape}, URPC~\citep{luo2021urpc}, and DTC~\citep{luo2021semi}; 
2) Noise perturbation: UA-MT~\citep{yu2019uamt}, MC-Net~\citep{wu2021semi}, MC-Net+~\citep{wu2022mutual}; 
3) CutMix perturbation: FixMatch~\citep{sohn2020fixmatch}, MT~\citep{tarvainen2017mean}, CPS~\citep{chen2021semi};
and 4) UMix and I-UMix perturbations: UCMT~\citep{shen2023co} and ABD~\citep{chi2024adaptive}.

\begin{table*}[!t]
\centering
\caption{Comparison with state-of-the-art methods on the LA dataset with 10\% $|\mathcal{D}|$ and 20\% $|\mathcal{D}|$ labeled data. Note that all methods adopt V-Net as the backbone.
The best and second-best results are highlighted in {\color{red}\textbf{red}} and {\color{blue}\textbf{blue}}, respectively. 
${}^*$ indicates statistically significant improvements ($p$-value $ < 0.05$ based on a paired t-test) of our AdaMix frameworks over ABD, MT, and UCMT, respectively.}
\resizebox{\textwidth}{!}{
\begin{tabular}{lllll|llllll}
\toprule[1pt]
\multirow{2}{*}{Method} & \multicolumn{4}{c|}{10\% $|\mathcal{D}|$} & \multicolumn{4}{c}{20\% $|\mathcal{D}|$}  & \multirow{2}{*}{Perturbation}   & \multirow{2}{*}{Framework}\\ \cline{2-9}
& DSC (\%) $\uparrow$ & Jaccard (\%) $\uparrow$ & 95HD $\downarrow$ & ASD $\downarrow$  & DSC (\%) $\uparrow$ & Jaccard (\%) $\uparrow$ & 95HD $\downarrow$ & ASD $\downarrow$  \\ \midrule
URPC~\citep{luo2021urpc} &    85.01       &    74.36     &    15.37       &    3.96       &    88.74       &    79.93     &    12.73       &   3.66  & Multi-Task  & Self-Training  \\
SASSNet~\citep{li2020shape}     &    87.32       &    77.72     &    9.62       &    2.55       &    89.54       &    81.24     &    8.24       &   2.20  
 & Multi-Task  & Self-Training\\
FixMatch~\citep{sohn2020fixmatch}   & 87.88  & 78.53  & 10.33 &  2.68 
 & 89.22  & 80.65  & 7.80  & 2.29  & CutMix  & Self-Training \\
ABD~\citep{chi2024adaptive}  &     88.84      &   80.01      &     9.88      &    2.13  &     90.94      &     83.45      &     5.65      & 1.86  & UMix + I-UMix  & Self-Training \\ 
AdaMix-ST (ours)   &     89.60${}^*$      &    81.37${}^*$     &     {\color{blue}\textbf{8.12}}${}^*$      &    {\color{blue}\textbf{2.01}}  &     91.40      &     84.24${}^*$      &     {\color{blue}\textbf{5.03}}${}^*$       &    {\color{blue}\textbf{1.51}}  & Adaptive Mix  & Self-Training  \\  \midrule 
 \midrule
UA-MT~\citep{yu2019uamt}     &    84.25       &    73.48     &    13.84      &    3.36      &    88.88       &    80.21     &    7.32      &    2.26  & Noise  & Mean-Teacher \\
MT~\citep{tarvainen2017mean}      &    86.15   &      76.16   &     11.37      &       3.60  &     89.81    &      81.85   &    6.08     &       1.96  & CutMix  & Mean-Teacher \\ 
AdaMix-MT (ours)   &     {\color{blue}\textbf{90.20}}${}^*$      &    {\color{blue}\textbf{81.98}}${}^*$     &     8.32${}^*$     &    2.04${}^*$  &     {\color{red}\textbf{91.87}}${}^*$     &     {\color{red}\textbf{85.36}}${}^*$      &    5.53${}^*$       &    1.65${}^*$   & Adaptive Mix  & Mean-Teacher \\ \midrule 
 \midrule
DTC~\citep{luo2021semi}     &    86.57     &      76.55   &     14.47     &      3.74   &    89.42     &      80.98   &     7.32     &      2.10  & Multi-Task  & Co-Training  \\  
MC-Net~\citep{wu2021semi}      &     87.71    &      78.31   &     9.36      &       2.18  &     90.34    &      82.48   &     6.00     &       1.77  & Noise  & Co-Training \\ 
MC-Net+~\citep{wu2022mutual}      &     88.96    &      80.25   &     7.93      &       1.86  &     91.07    &      83.67   &     5.84     &       1.67  
 & Noise  & Co-Training \\  
CPS~\citep{chen2021semi}      &    86.23    &      76.22   &     11.68      &       3.65  &     88.72    &      80.01   &    7.49     &       1.91  & CutMix  & Co-Training \\  
UCMT~\citep{shen2023co}   &     88.13      &    79.18      &     9.14      &    3.06  &     90.41      &     82.54      &     6.31      & 1.70  & UMix  & Co-Training\\ 
AdaMix-CT (ours)  &     {\color{red}\textbf{90.69}}${}^*$     &    {\color{red}\textbf{83.21}}${}^*$     &    {\color{red}\textbf{5.37}}${}^*$    &    {\color{red}\textbf{1.57}}${}^*$  &     {\color{blue}\textbf{91.58}}${}^*$      &     {\color{blue}\textbf{85.16}}${}^*$      &     {\color{red}\textbf{4.95}}${}^*$      &    {\color{red}\textbf{1.44}}  & Adaptive Mix  & Co-Training \\
\bottomrule[1pt]
\end{tabular}}
\label{Tab:la}
\end{table*}

\begin{table*}[!t]
\centering
\caption{Comparison with state-of-the-art methods on the ISIC dataset with 5\% $|\mathcal{D}|$ and 10\% $|\mathcal{D}|$ labeled data. 
Note that all methods adopt U-Net as the backbone.
The best and second-best results are highlighted in {\color{red}\textbf{red}} and {\color{blue}\textbf{blue}}, respectively.
${}^*$ indicates statistically significant improvements ($p$-value $ < 0.05$ based on a paired t-test) of our AdaMix frameworks over ABD, MT, and UCMT, respectively.}
\resizebox{\textwidth}{!}{
\begin{tabular}{lllll|llllll}
\toprule[1pt]
\multirow{2}{*}{Method} & \multicolumn{4}{c|}{5\% $|\mathcal{D}|$} & \multicolumn{4}{c}{10\% $|\mathcal{D}|$}  &\multirow{2}{*}{Perturbation \& Framework} \\ \cline{2-9}
& DSC (\%) $\uparrow$ & Jaccard (\%) $\uparrow$ & 95HD $\downarrow$ & ASD $\downarrow$  & DSC (\%) $\uparrow$ & Jaccard (\%) $\uparrow$ & 95HD $\downarrow$ & ASD $\downarrow$  \\ \midrule
FixMatch~\citep{sohn2020fixmatch}   & 84.16  & 75.67  & 12.29  &  2.25  & 
 85.42   & 77.35  & 10.43  & 2.07   & CutMix  & Self-Training  \\
ABD~\citep{chi2024adaptive}  & 84.68  &  76.56  &  12.41  &  3.72   & 86.27  & 78.38  &  10.17  &  2.19  & UMix + I-UMix  & Self-Training \\  
AdaMix-ST (ours)   &     {\color{blue}\textbf{85.78}}${}^*$      &  {\color{blue}\textbf{77.78}}${}^*$      &    {\color{blue}\textbf{11.39}}${}^*$      &   {\color{blue}\textbf{2.36}}${}^*$   &     {\color{red}\textbf{87.60}}${}^*$      &    {\color{red}\textbf{79.47}}${}^*$      &    {\color{red}\textbf{9.18}}${}^*$      &   {\color{red}\textbf{1.74}}${}^*$    & Adaptive Mix  & Self-Training  \\ \midrule \midrule
UA-MT~\citep{yu2019uamt} &    81.65       &    72.47   &    17.10     &    4.86      &    83.18       &    74.33     &    13.89     &    2.79   & Noise  & Mean-Teacher \\
MT~\citep{tarvainen2017mean}      &    83.05   &     74.36   &    15.51      &       4.40  &     85.06    &      76.68   &    11.49     &       2.51  & CutMix  & Mean-Teacher  \\  
AdaMix-MT (ours)  &     85.48${}^*$      &   77.27${}^*$      &     11.48${}^*$      &   2.65${}^*$ &     86.83${}^*$      &  78.97${}^*$     &    10.09${}^*$      &   2.03   & Adaptive Mix  & Mean-Teacher  \\ \midrule \midrule
CPS~\citep{chen2021semi}  &    83.03   &     74.06   &    14.67     &       3.80  &     84.99    &      76.79   &    11.13     &       2.20  & CutMix  & Co-Training  \\ 
UCMT~\citep{shen2023co}   &    84.30   &     76.03   &    11.52    &       2.65  &     86.57    &      78.89   &    9.53     &       2.06  & UMix  & Co-Training \\
AdaMix-CT (ours)   &    {\color{red}\textbf{85.94}}${}^*$      &    {\color{red}\textbf{77.81}}${}^*$     &     {\color{red}\textbf{11.02}}${}^*$      &   {\color{red}\textbf{2.31}}${}^*$   &     {\color{blue}\textbf{87.09}}${}^*$     &  {\color{blue}\textbf{79.05}}${}^*$      &    {\color{blue}\textbf{9.24}}     &   {\color{blue}\textbf{1.91}}   & Adaptive Mix  & Co-Training   \\
\bottomrule[1pt] 
\end{tabular}}
\label{Tab:isic}
\end{table*}

\subsubsection{Results on ACDC}
Table~\ref{Tab:acdc} shows the results of our three AdaMix frameworks (\emph{i.e.}, AdaMix-ST, AdaMix-MT, and AdaMix-CT) and the compared methods on the ACDC dataset with 10\% and 20\% labeled data. All the methods are evaluated on the average segmentation performance of left ventricular (LV) cavity, right ventricular (RV) cavity, and myocardium (MYO). 
The proposed AdaMix frameworks consistently outperform other approaches among all the metrics.
Specifically, AdaMix-CT achieves the best performance of 89.19\% and 89.83\% in terms of DSC under 10\% and 20\% labeled data, respectively. Concerning the boundary-sensitive metrics 95HD and ASD, with the help of AdaMix, AdaMix-CT obtains the best values of 2.46 and 0.74 for the 10\% labeled data. 
This implies that AdaMix can facilitate the model to delineate segmentation boundaries better. 
On the one hand, multi-task consistency-based approaches perform poorly in low data regions, \emph{i.e.}, 10\% labeled data, indicating that the consistency regularization effects brought by auxiliary tasks, such as signed distance maps in SASSNet~\citep{li2020shape} and DTC~\citep{luo2021semi}, or deep supervision in URPC~\citep{luo2021urpc}, are limited. 
On the other hand, due to uncontrollable perturbation strength, the models equipped with either random noise or CutMix show a performance gap compared to our AdaMix framework.
For instance, compared with the representative co-training-based model, MCNet+, our AdaMix-CT shows a performance gain of 2.09\% and 1.32\% in terms of DSC under 10\% and 20\% labeled data, respectively.
Moreover, our AdaMix frameworks significantly outperform the SSL methods with fixed mix-up rules, \emph{i.e.}, UCMT~\citep{shen2023co} and ABD~\citep{chi2024adaptive}. For example, compared with UCMT, AdaMix-CT obtains relative improvements of 2.62\% in Dice similarity coefficient and 48.25\% in average surface distance on the ACDC dataset with 10\% labeled data.
These results demonstrate the superiority of Adaptive Mix for consistency regularization in semi-supervised medical image segmentation.

\subsubsection{Results on LA}
We further evaluated the proposed method on a 3D image segmentation scenario for left atrial segmentation.
Table~\ref{Tab:la} reports the results on the LA dataset with 10\% (8 labeled samples) and 20\% (16 labeled samples) labeled data.
In general, our AdaMix frameworks substantially outperform all the compared methods in terms of the four metrics. 
Specifically, the AdaMix frameworks show competitive performance in surface-related metrics, indicating that AdaMix, through adaptively adjusting perturbation degrees, generates appropriate perturbed images, thereby facilitating the frameworks' advantages in 3D boundary segmentation.
For example, AdaMix-CT obtains the best ASD values of 1.57 and 1.44 under 10\% and 20\% labeled data, respectively.
These results prove the advantages of our method for semi-supervised volumetric medical image segmentation.

\begin{figure*}[!t]
\centerline{\includegraphics[width=0.85\linewidth]{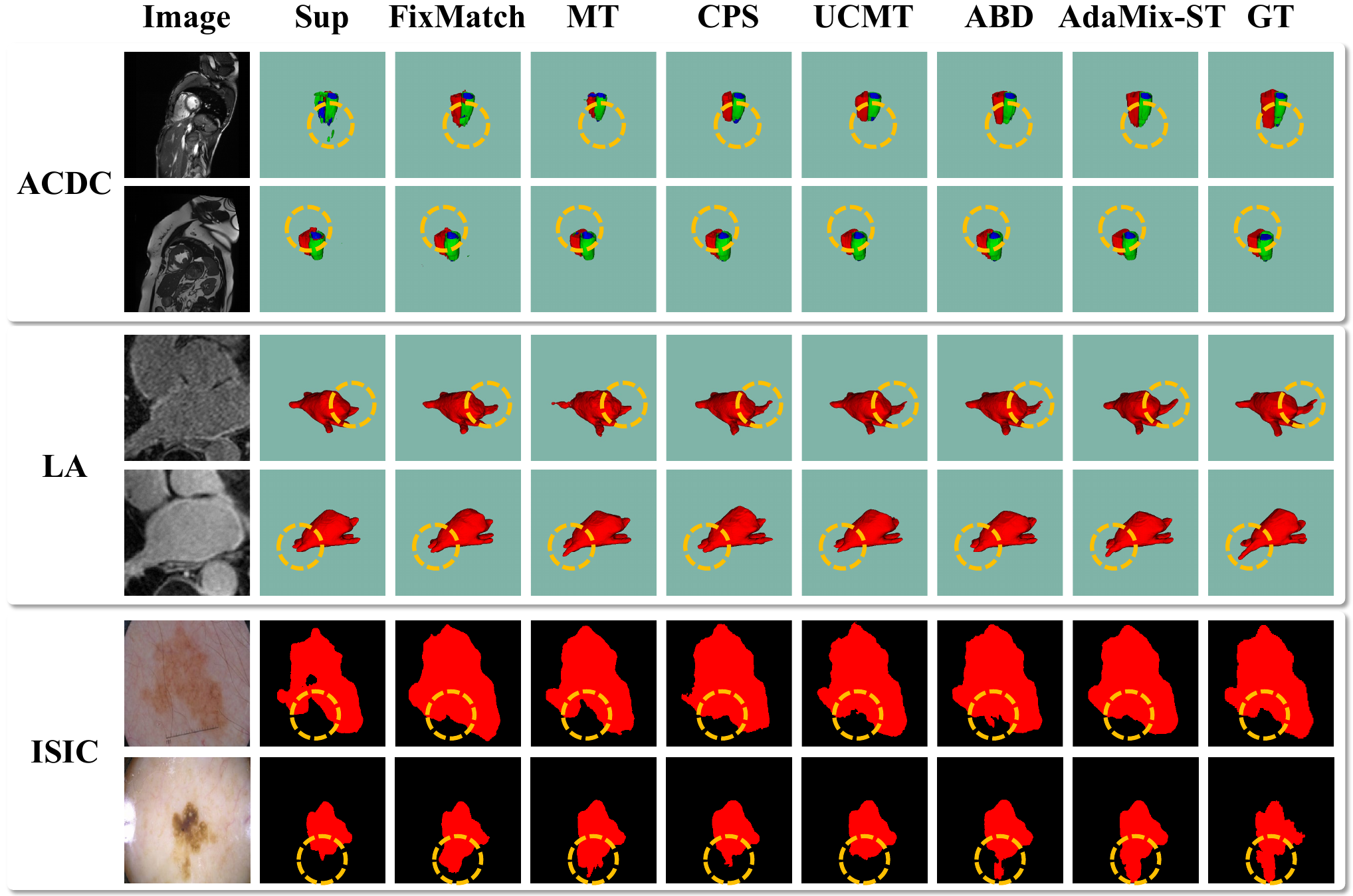}}
\caption{Qualitative examples on the ACDC, ISIC, and LA datasets. The yellow dash circles highlight some segmentation regions. 
Sup: the supervised baseline, MT: Mean-Teacher, CPS: Cross Pseudo Supervision, and GT: Ground Truth.
}
\label{fig:qualitative}
\end{figure*}

\subsubsection{Results on ISIC}
Unlike previous datasets for cardiac structure segmentation from MRI images, the ISIC dataset involves skin lesion segmentation from dermoscopy images, which includes heterogeneous segmentation objects.
The results are reported in Table~\ref{Tab:isic}. 
It can be observed that 
in comparison with the methods utilizing CutMix~\citep{yun2019cutmix} perturbation (\emph{e.g.}, FixMatch~\citep{sohn2020fixmatch}, MT~\citep{tarvainen2017mean}, and CPS~\citep{chen2021semi}) or UMix~\citep{shen2023co} perturbation (\emph{e.g.}, UCMT~\citep{shen2023co} and ABD~\citep{chen2021semi}), our AdaMix frameworks demonstrate compelling performance.
For instance, compared with UCMT~\citep{shen2023co} and ABD~\citep{chi2024adaptive}, AdaMix-CT shows performance gains of 1.64\% and 1.26\% in terms of DSC respectively under 5\% labeled data. 
Remarkably, in light of boundary-sensitive metrics, \emph{i.e.}, 95HD and ASD, the proposed methods obtain the best scores due to sufficient training on the challenging target locations and boundaries in the perturbed images provided by AdaMix.
These results further demonstrate the superiority of our method for heterogeneous lesion segmentation.

\textbf{Remark.}
One can observe from Table~\ref{Tab:acdc}, Table~\ref{Tab:la}, and Table~\ref{Tab:isic} that the results obtained by the three AdaMix frameworks are comparable, validating the flexibility of AdaMix to act as an effective plug-in-play module in SSL paradigms.
In contrast, the methods with different perturbation strategies show significant performance discrepancies. For example, MT and AdaMix-MT exhibit a performance gap of 4.05\% in terms of DSC on the LA dataset with 10\% labeled data, while the margin between MT and CPS is only 0.08\%.
The results suggest that perturbation strategies are more critical than learning paradigms for enhancing the models' performance in consistency regularization-based methods. 
This phenomenon can be attributed to the data-driven nature: DNNs are data-driven, making the quality of training data more crucial for accurately fitting the real data distribution in the underlying task, where the quality of training data can be considered as the degree to which the training data distribution represents the real data distribution.

\subsubsection{Qualitative Results}
\label{subsec:qualitative_results}
As shown in Fig.~\ref{fig:qualitative}, we visualize some segmentation examples on the ACDC, LA, and ISIC datasets. 
Specifically, the supervised baseline is prone to under-segment some lesion regions due to the limited number of labeled data. 
FixMatch~\citep{sohn2020fixmatch}, MT~\citep{tarvainen2017mean}, and CPS~\citep{chen2021semi} (all using CutMix~\citep{yun2019cutmix} as the strong perturbation) incorrectly classify many boundary pixels, mainly due to noise introduced by random perturbations during training.
Similarly, UCMT~\citep{shen2023co} and ABD~\citep{chi2024adaptive}, which use UMix~\citep{shen2023co} as the strong perturbation, also exhibit inferior performance on segmentation boundaries. These results can be attributed to the limited effectiveness of consistency regularization.
In contrast, our AdaMix-ST exhibits smoother and clearer boundaries compared to other methods. The finer segmentation results can be ascribed to the AdaMix algorithm, which adaptively generates perturbed images in a self-paced learning manner.

\begin{table*}[!t]
\centering
\caption{Ablation study of the AdaMix-ST framework on the ACDC \textit{val} set under 10\% labeled data. Sup: the supervised baseline. 
Note that without the self-paced mask, AdaMix is performed according to a fixed rule, \emph{i.e.}, replacing low-confidence patches with high-confidence ones. The best results are highlighted in \textbf{bold}.
}
\resizebox{\textwidth}{!}{
\begin{tabular}{c|ccc|cccc}
\toprule[1pt]
\multirow{2}{*}{Method} & \multirow{2}{*}{Strong-Weak Supervision}
& \multicolumn{2}{c|}{AdaMix}  & \multicolumn{4}{c}{10\%} \\
& & Self-Paced Mask & Self-Paced Weight
& DSC (\%) $\uparrow$ & Jaccard (\%) $\uparrow$ & 95HD $\downarrow$ & ASD $\downarrow$ \\
\midrule
Sup 	&    &   &   & 83.85  & 73.67  & 6.54  & 2.97\\
SSL-ST	&  $\surd$  &   &    &    85.59       &    76.18  & 5.97  &  2.88 \\ 
AdaMix-ST (V1) 	&  $\surd$  &  $\surd$ &   &     87.67     &    78.50   &    6.64   &    1.82 \\ 
AdaMix-ST (V2) 	&  $\surd$  &    & $\surd$  &    87.84       &    78.94   &    4.63      &    1.71 \\ 
AdaMix-ST (V3) 	&  $\surd$  & $\surd$  &  $\surd$  &     \textbf{88.86}  & \textbf{81.03}  & \textbf{4.46}    &   \textbf{1.24} \\
\bottomrule[1pt]
\end{tabular}}
\label{Tab:ablation}
\end{table*}

\begin{figure*}[!t]
\centerline{\includegraphics[width=\textwidth]{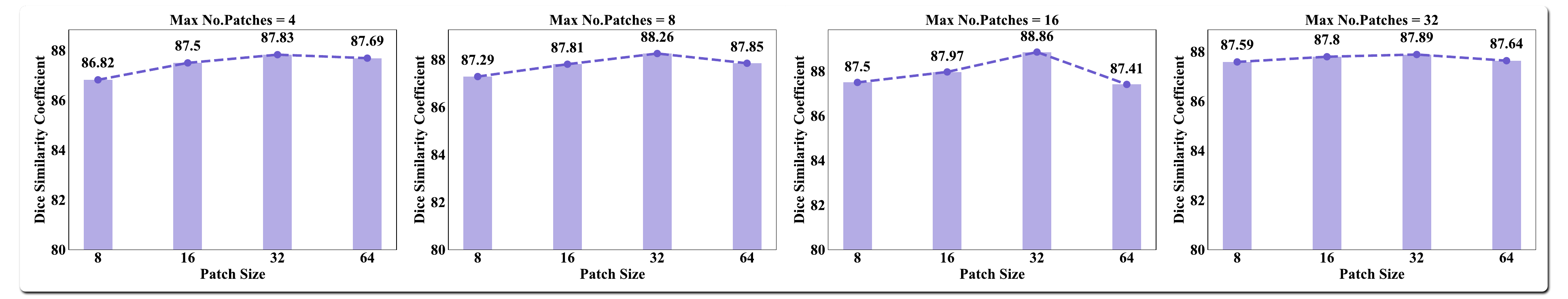}}
\caption{Investigation of the effect of patch size $S$ and maximum No. patches $K$ for AdaMix on the ACDC \textit{val} set with 10\% labeled data.}
\label{fig:ab_patchsize_nopatch_acdc}
\end{figure*}

\begin{figure}[!t]
\centerline{\includegraphics[width=\linewidth]{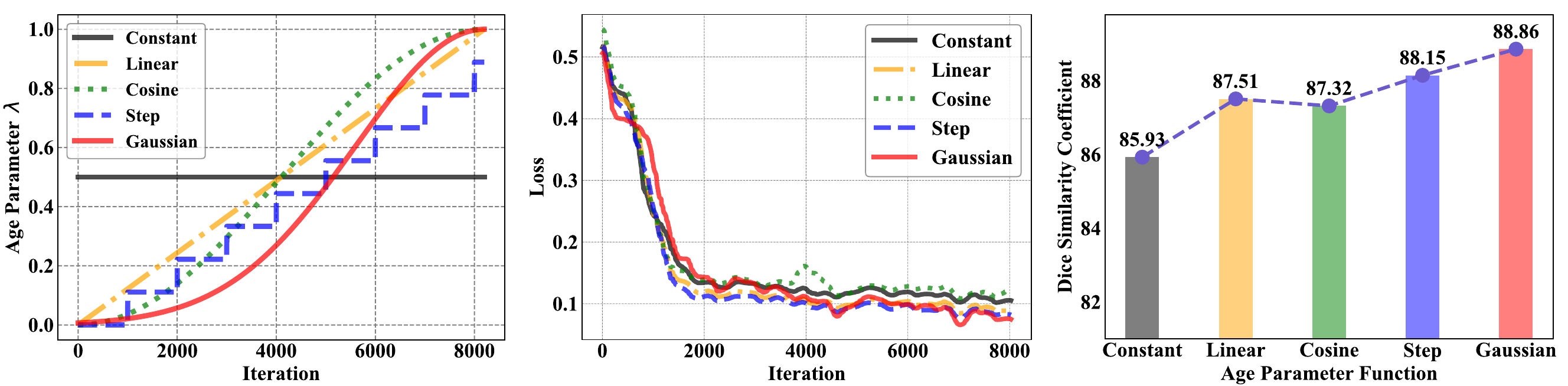}}
\caption{Analysis of the effect of age parameter functions for AdaMix on the ACDC dataset with 10\% labeled data. Left: the age parameter function curves; middle: the training loss curves of models with different age parameter functions; right: the corresponding segmentation performance on the ACDC \textit{val} set.}
\label{fig:ab_aga_para_func_acdc}
\end{figure}

\subsection{Ablation Study}
\label{subsec:ablation_study}
In the following, we conducted extensive ablation experiments to investigate the contribution of each component of the proposed method in detail [Table~\ref{Tab:ablation}]. Specifically, we took our AdaMix-ST as an example and examined the effects of AdaMix by considering three versions: 1) AdaMix-ST (V1): AdaMix-ST without the self-paced weight, 2) AdaMix-ST (V2): AdaMix-ST without the self-paced mask, and 3) AdaMix-ST (V3): the final method. Meanwhile, we explored in-depth the influence of hyperparameters, \emph{i.e.}, the maximum number of mix-up patches $K$ and the patch size $S$ [Fig.~\ref{fig:ab_patchsize_nopatch_acdc}], as well as the age parameter $\lambda$ [Fig.~\ref{fig:ab_aga_para_func_acdc}].
Moreover, we also provide a theoretical and experimental analysis of the effect of AdaMix on training convergence [Fig.~\ref{fig:convergence_analysis}], and visualize several AdaMix examples during training [Fig.~\ref{fig:adamix_procedure_vis}].

\subsubsection{Improvement over the Supervised Baseline}
Compared with the supervised baseline, our final method (\emph{i.e.}, AdaMix-ST (V3)) achieves considerable relative improvements of 5.97\%, 9.99\%, 31.80\%, and 58.25\%, in terms of DSC, Jaccard, 95HD, and ASD, respectively. 
These results demonstrate the effectiveness of our AdaMix frameworks for semi-supervised medical image segmentation. 

\subsubsection{Effectiveness of Each Component}
Table~\ref{Tab:ablation} shows a trend that the segmentation performance improves when we gradually incorporate the proposed components into our method.

\paragraph{Effectiveness of Self-Paced Mask}
It can be observed that when a self-paced mask is introduced into the model to determine the mix-up rule based on the model's learning state, the results of AdaMix-ST (V1) are higher than those of SSL-ST, \emph{e.g.}, with 2.08\% and 1.06 improvement in DSC and ASD, respectively.
\paragraph{Effectiveness of Self-Paced Weight}
In the comparison between AdaMix-ST (V2) and SSL-ST, it can be seen that introducing the self-paced weight into our model to determine the number of mix-up patches results in a 2.25\% improvement in DSC and a 1.17 improvement in terms of ASD.

We can conclude that the self-paced curriculum for determining the mix-up rule and the number of patches is beneficial for semi-supervised segmentation. Building upon the self-paced curriculum, our final model, AdaMix-ST (V3), shows significant improvements over the semi-supervised baseline in both global segmentation metrics (DSC and Jaccard) and boundary-sensitive metrics (95HD and ASD), achieving the best performance with 88.86\% DSC, 81.03\% Jaccard, 4.46 95HD, and 1.24 ASD.

\begin{figure*}[!t]
\centerline{\includegraphics[width=.75\linewidth]{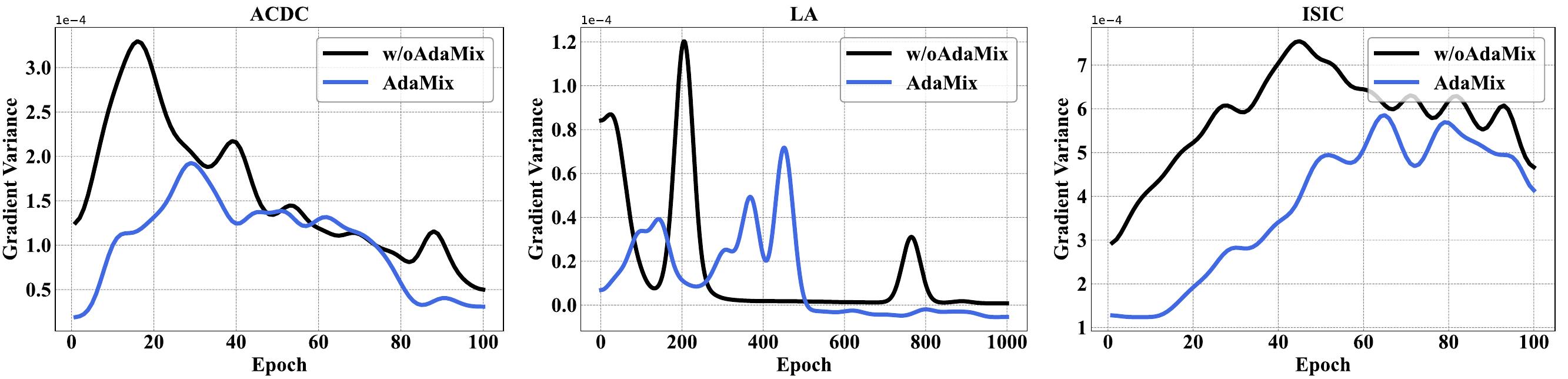}}
\caption{Gradient variances of models trained with and without AdaMix on the ACDC, LA, and ISIC datasets. 
Each point on the curves represents the norm of the batch-wise gradient variances for the entire training set.
} 
\label{fig:gradient_variances}
\end{figure*}

\begin{figure*}[!t]
\centerline{\includegraphics[width=.75\linewidth]{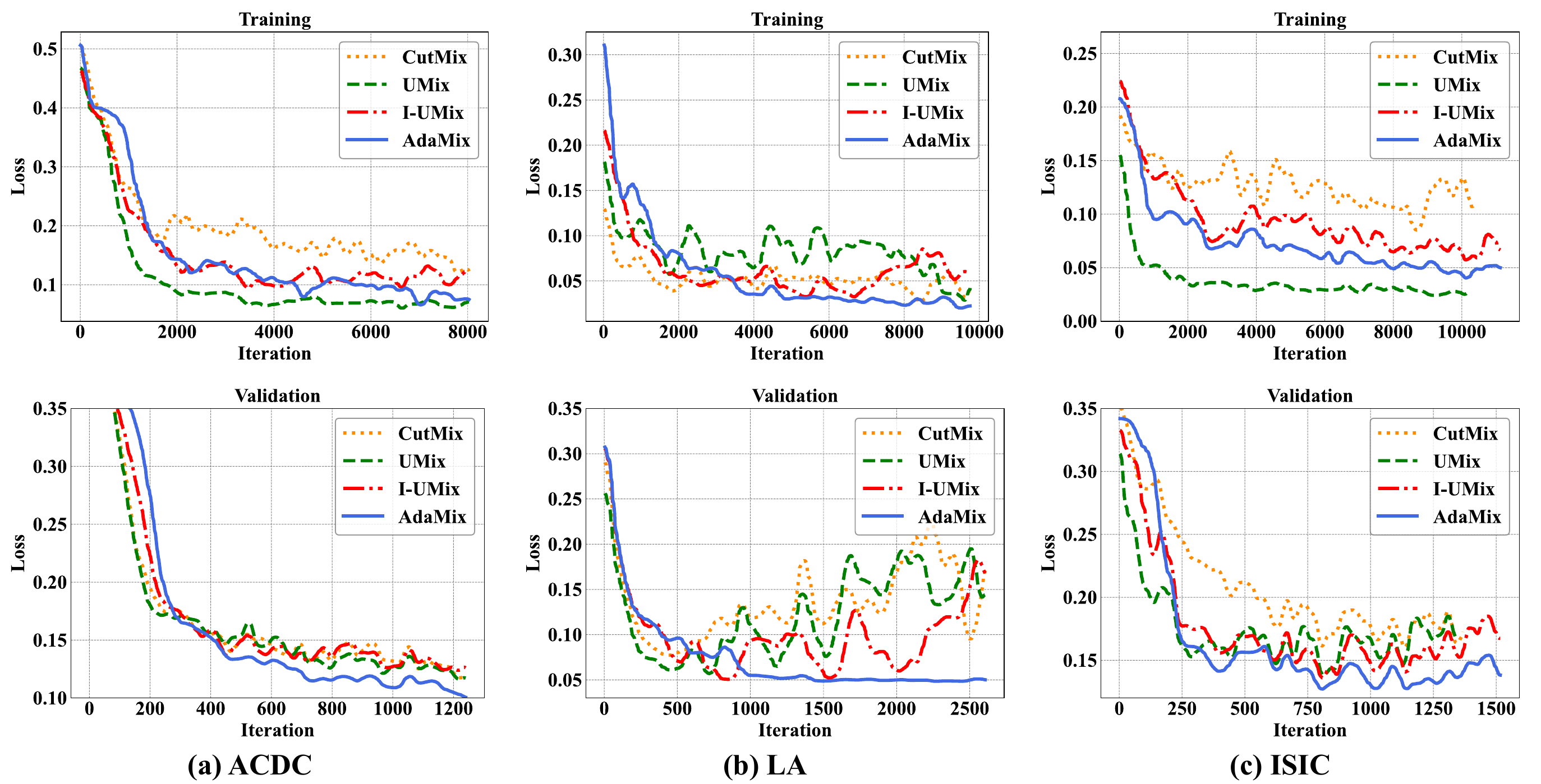}}
\caption{Training and validation losses of CutMix, UMix, I-UMix, and the proposed AdaMix on the (a) ACDC, (b) LA, and (c) ISIC datasets, respectively.} 
\label{fig:convergence_analysis}
\end{figure*}

\subsubsection{Analysis of Hyperparameters}
\label{subsubsec:hyperparameters}
We investigated the influence of the mix-up patch size $S$ and the maximum number of mix-up patches $K$ on AdaMix, using the ACDC \textit{val} set with 10\% labeled data.
The results are shown in Fig.~\ref{fig:ab_patchsize_nopatch_acdc}, where each subplot illustrates how the DSC changes with varying patch sizes when the maximum number of mix-up patches is fixed.
In general, AdaMix is not sensitive to changes in $K$, whereas variations in $S$ cause slight performance fluctuations.
Specifically, increasing the patch size marginally improves performance, with the highest DSC score achieved when $S = 32$.
The main reasons behind this phenomenon are: 1) the proposed self-paced learning strategy can adaptively adjust the number of mix-up patches based on the model's learning state, where $K$ only determines the maximum number of mix-up patches, and 2) patches that are too small or too large inevitably introduce noise to the images; small patches fail to provide complete anatomical information, while large patches disrupt the anatomical structure of original images.
Moreover, it is surprising that the impact of the mix-up patch size $S$ on the variation in performance is more trivial when the maximum number of mix-up patches $K=32$. We believe that with a large number of mix-up patches, AdaMix can control a wider range of mix-up degrees. Consequently, it exhibits less sensitivity to hyperparameters, making the algorithm more robust.

\subsubsection{Analysis of Age Parameter Functions}
We explored the impact of different age parameter functions for AdaMix on the ACDC dataset with 10\% labeled data.
As shown in Fig.~\ref{fig:ab_aga_para_func_acdc}, while different function shapes lead to variations in the adaptive perturbation procedure, all models with these functions converge to minima during training.
However, these variations in adaptive perturbation impact the segmentation performance of the trained models.
Notably, the age parameter $\lambda$ increasing with a Gaussian ramp-up function contributes to the highest DSC score, compared with the other functions. 
These results suggest that the Gaussian function curve aligns better with the model's state in the training process and validates our hypothesis that the model's performance progressively improves during training.

\begin{figure*}[!t]
\centerline{\includegraphics[width=0.85\linewidth]{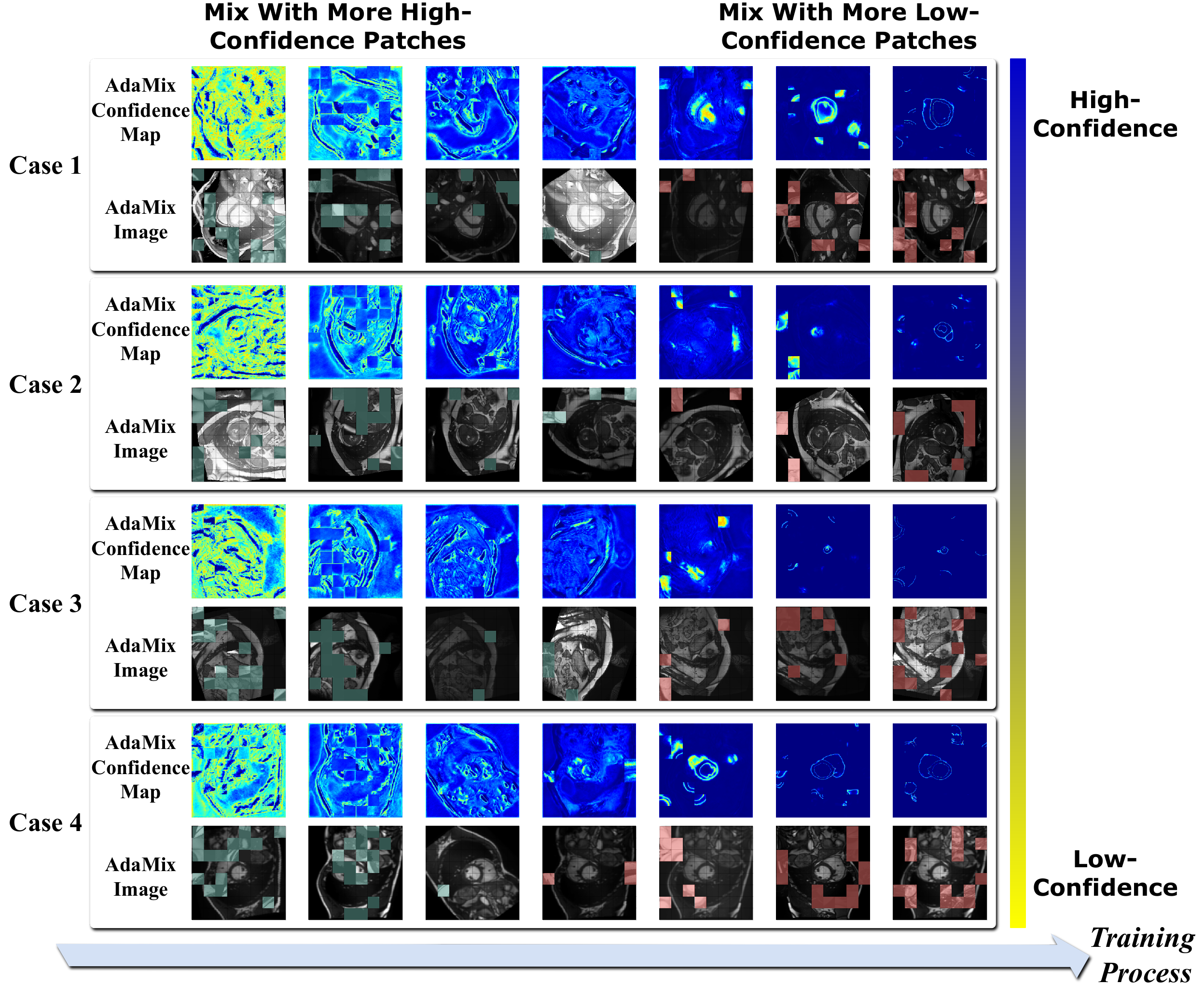}}
\caption{Visualization of the AdaMix procedure during training: AdaMix initially generates perturbed samples containing more high-confidence regions, and then gradually incorporates low-confidence regions into the perturbed images as training progresses ({\color{teal}\textbf{green}} regions: the high-confidence patches from auxiliary images; {\color{purple}\textbf{red}} regions: the low-confidence patches from auxiliary images). According to the confidence maps, high-confidence regions generally appear in the background, while low-confidence ones tend to occur at the boundaries of the segmentation targets.
}
\label{fig:adamix_procedure_vis}
\end{figure*}

\subsubsection{Convergence Analysis}
\label{subsubsec:convergence_analysis}
Based on the optimization theory for large-scale machine learning~\citep{bottou2018optimization}, we first provide a theoretical analysis that the AdaMix algorithm can generate perturbed samples to reduce gradient variance for facilitating faster convergence of the segmentation model training. 

Let $J(\theta)$ be the empirical risk over the dataset $\mathcal{D}$ for the segmentation loss function $\mathcal{L}(\theta)$ with the model $f(\cdot; \theta)$. 
We assume that the loss function is $\ell$-gradient Lipschitz and that the gradient estimate has bounded variance.
Based on the \textit{$\ell$-smooth ($\ell$-gradient Lipschitz)} and the \textit{bounded gradient variance} assumptions, we have the following convergence theorem.
\begin{theorem}
    Convergence theorem. Using the stochastic gradient descent with a learning rate $\eta_t = \frac{\eta_0}{t}$ at the $t^{th}$ iteration ($\eta_0$ refers to the initial learning rate) and the total optimization steps $T$, the convergence rate satisfies:
    \begin{equation}
        \frac{1}{T}\sum_{t=1}^T\mathbb{E}[\Vert J(\theta_t) \Vert^2] \leq \frac{2(J(\theta_0) - J^*))}{\eta_0\sqrt{T}} + \frac{\eta_0\ell\sigma^2}{\sqrt{T}}
        \nonumber
    \end{equation}
    where the term $\frac{\eta_0\ell\sigma^2}{\sqrt{T}}$ dominates the bound for large $T$. 
\end{theorem}
This theorem implies that a smaller $\sigma^2$ ensures faster convergence to a stationary point.  
Since AdaMix performs adaptive patch-wise mix-up between two images and its loss function is the weighted average over these regions, the resulting loss function for AdaMix samples still satisfies $\ell$-smooth, and its Lipschitz constant remains $\ell$.
Mix-up operations between different images generate the synthetic samples, which lie in the neighborhood of the original images, leading to a smoother estimate of the gradient, i.e. a lower gradient variance $\sigma^2_x \leq \sigma^2$ ~\citep{zhang2017mixup,yun2019cutmix}.

We conducted experimental validation to support this argument. To this end, we compared the gradient variances of models trained with and without AdaMix on the ACDC, LA, and ISIC datasets.
As shown in Fig.~\ref{fig:gradient_variances}, the model trained with AdaMix has lower gradient variances, indicating a more stable training process. Notably, AdaMix maintains consistently smaller gradient variances throughout the training process, except for some fluctuations observed during the mid-training phase on the LA dataset.
This result corroborates our theoretical analysis, confirming that the AdaMix algorithm reduces gradient variance.

We visualize the training and validation loss curves of CutMix, UMix, I-UMix, and our AdaMix on the ACDC, LA, and ISIC datasets in Fig.~\ref{fig:convergence_analysis}, respectively. 
According to the training loss curves, these methods generally converge to their minima. 
It can be observed that both CutMix and I-UMix exhibit more dramatic oscillations during the training process, especially on the ISIC dataset. This is because the random mixing operation or the mixing with low-confidence regions tends to generate perturbed samples that deviate from the model's current state.
Meanwhile, UMix leads to a faster decrease in training loss but with a relatively inferior generalization performance on the validation sets, as mixing with high-confidence regions produces easier samples.
In contrast, AdaMix, with its adaptive perturbation strength, ensures a more stable training process.
On the other hand, the lower validation loss of AdaMix indicates that it leads to a more robust trained model, \emph{i.e.}, against overfitting and with better generalization capability.
In summary, through the empirical experiments, we demonstrate that the AdaMix algorithm reduces gradient variance, leading to faster training convergence.

\subsubsection{Analysis of Adaptive Mix Samples}
\label{subsubsec:adamix_samples}
AdaMix provides perturbed image examples with gradually increasing sample complexity, facilitating the effective consistency regularization on unlabeled data.  
As shown in Fig.~\ref{fig:adamix_procedure_vis}, taking four cases from the ACDC dataset as an example, we illustrate the AdaMix procedure during training.
Overall, we can observe that high-confidence regions typically appear in the background, while low-confidence ones tend to appear at the boundaries of the segmentation targets. In medical image segmentation tasks, the background region contains simple patterns, while the object boundaries are considered complex ones~\citep{yu2019uamt,shen2023co}.
Deep neural networks learn simple patterns first~\citep{arpit2017closer}.
In the initial stage of training, considering the model's inferior capability, AdaMix provides perturbed samples with more high-confidence regions (\emph{i.e.}, more background regions), from which the model can generate relatively accurate predictions. 
As training progresses, the model's performance improves; AdaMix gradually incorporates low-confidence patches that occur at challenging target boundaries into the perturbed images, providing more complex samples for training; forcing the model to focus on these regions enhances its discriminative capability in the later training stages.
In light of the above, AdaMix can adaptively adjust the perturbation strength based on the model's learning state, achieving more efficient consistency regularization.

\section{Conclusion}
\label{sec:conclusion}
This paper explores adaptive mix-up techniques for semi-supervised medical image segmentation. 
We propose a novel Adaptive Mix algorithm (AdaMix) that features a self-paced curriculum, which initially provides relatively simple mixed samples and then gradually increases the difficulty of the mixed images by adaptively controlling the mix-up strength based on the model's learning state. 
AdaMix can be seamlessly integrated as a plug-and-play module into self-training, mean-teacher, and co-training semi-supervised learning paradigms, achieving superior performance. Experiments on three public medical image datasets validate the effectiveness of the proposed method and demonstrate its superiority over state-of-the-art approaches.
Moreover, we reveal that incorporating more high-confidence patches into images generates easier samples, which benefits the segmentation model during the initial training stage; in contrast, replacing high-confidence regions with low-confidence ones produces harder examples that are effective for the later training stage. Since low-confidence regions often appear in segmentation boundaries, forcing the model to learn from these regions helps enhance its discriminative capability.
Our analysis also demonstrates that, in strong-weak pseudo supervision for semi-supervised learning, perturbation strategies are more critical than learning paradigms.

\section*{CRediT authorship contribution statement}
\textbf{Zhiqiang Shen}: Writing – original draft, Methodology, Visualization, Validation, Data curation, Conceptualization.
\textbf{Peng Cao}: Writing – review \& editing, Supervision, Project administration, Investigation, Funding acquisition, Formal analysis, Conceptualization.
\textbf{Junming Su}: Data curation, Visualization, Validation. 
\textbf{Jinzhu Yang}: Investigation, Funding acquisition.
\textbf{Osmar R. Zaiane}: Writing – review \& editing, Formal analysis.

\section*{Declaration of competing interest}
The authors declare that they have no known competing financial interests or personal relationships that could have appeared to influence the work reported in this paper.

\section*{Acknowledgments}
This research was supported by the National Natural Science Foundation of China (No.62076059), the Science and Technology Joint Project of Liaoning Province (2023JH2/101700367), the Fundamental Research Funds for the Central Universities (N2424010-7), and the China Scholarship Council (202406080040).

\bibliographystyle{elsarticle-harv}
\bibliography{main}

\end{document}